\documentclass[10pt,twocolumn,letterpaper]{article}

\usepackage{iccv}
\usepackage{times}
\usepackage{epsfig}
\usepackage{graphicx}
\usepackage{amsmath}
\usepackage{amssymb}

\usepackage[breaklinks=true,bookmarks=false]{hyperref}

\iccvfinalcopy 

\usepackage{times}
\usepackage{epsfig}
\usepackage{graphicx}
\usepackage{amsmath}
\usepackage{amssymb}
\usepackage{footnote}
\usepackage{times}
\usepackage{epsfig}
\usepackage{graphicx}
\usepackage{amsmath}
\usepackage{amssymb}
\usepackage{subcaption}

\usepackage{tablefootnote}

\usepackage{booktabs}
\usepackage{multirow}
\usepackage{pifont}

\newcommand{\Paragraph}[1]{\vspace{1mm} \noindent \textbf{#1} \hspace{0mm}}

\usepackage{iccv}
\usepackage{times}
\usepackage{epsfig}
\usepackage{graphicx}
\usepackage{amsmath}
\usepackage{amssymb}

\def\iccvPaperID{5910} 

\ificcvfinal\fi

\begin{document}

\title{MiniVLM: A Smaller and Faster Vision-Language Model}

\author{
Jianfeng Wang \quad
Xiaowei Hu \quad
Pengchuan Zhang \quad
Xiujun Li \quad
Lijuan Wang \quad \\
Lei Zhang \quad
Jianfeng Gao \quad
Zicheng Liu \\
Microsoft \\
{\tt\small \{jianfw,xiaowh,penzhan,xiul,lijuanw,leizhang,jfgao,zliu\}@microsoft.com}
}

\author{
Jianfeng Wang \quad
Xiaowei Hu \quad
Pengchuan Zhang \quad
Xiujun Li \quad
Lijuan Wang \quad \\
Lei Zhang \quad
Jianfeng Gao \quad
Zicheng Liu \\
Microsoft \\
{\tt\small \{jianfw,xiaowh,penzhan,xiul,lijuanw,leizhang,jfgao,zliu\}@microsoft.com}
}

\maketitle
\ificcvfinal\thispagestyle{empty}\fi

\begin{abstract}
Recent vision-language (VL) studies have shown remarkable progress by learning generic representations from massive image-text pairs with transformer models. While existing research has focused on achieving high accuracy with large pre-trained models, building a lightweight model is of great value in practice but is less explored. In this paper, we propose a smaller and faster VL model, MiniVLM, which can be finetuned with good performance on various downstream tasks like its larger counterpart. MiniVLM consists of two modules, a vision feature extractor and a transformer-based vision-language fusion module. We design a Two-stage Efficient feature Extractor (TEE) inspired by the one-stage EfficientDet~\cite{TanPL20} network to reduce the cost of visual feature extraction by $99\%$, compared to a baseline model. We adopt the MiniLM~\cite{abs-2002-10957} structure to reduce the computation cost of the transformer module after comparing different compact BERT models. In addition, we improve the MiniVLM pre-training by adding $7M$ Open Images data, which are pseudo-labeled
by a state-of-the-art captioning model. We also pre-train with high-quality image tags obtained from a strong tagging model to enhance cross-modality alignment. The large models are used offline without adding any overhead in fine-tuning and inference. With the above design choices, our MiniVLM reduces the model size by $73\%$ and the FLOPs by $99\%$ while maintaining $94-97\%$ accuracy on multiple VL tasks. We hope that MiniVLM helps ease the use of the state-of-the-art VL research for on-the-edge applications.

\end{abstract}

\section{Introduction}
\begin{figure}
	\centering
	\begin{tabular}{@{}c@{}c@{}}
	\includegraphics[width=0.47\linewidth]{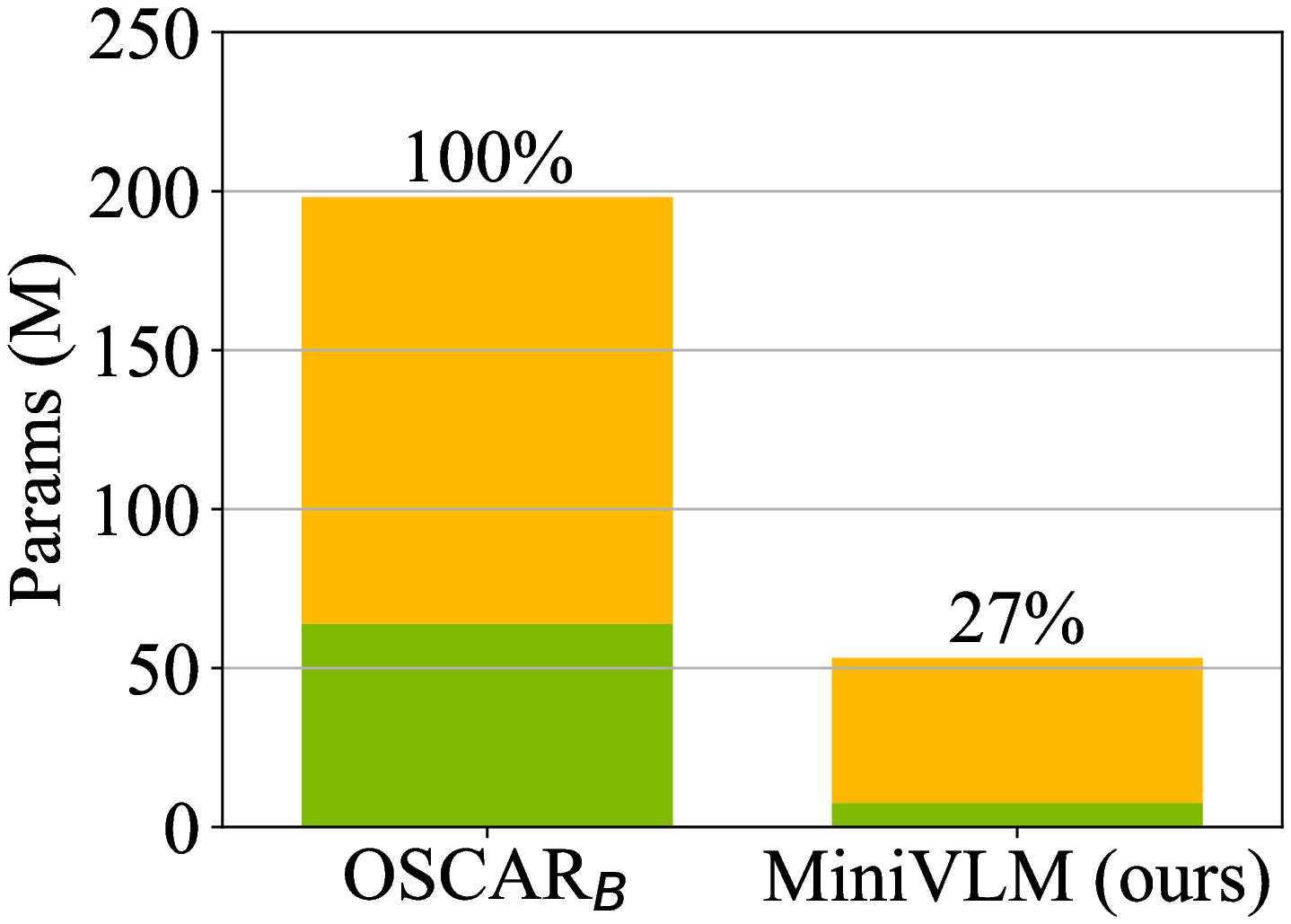} &
	\includegraphics[width=0.47\linewidth]{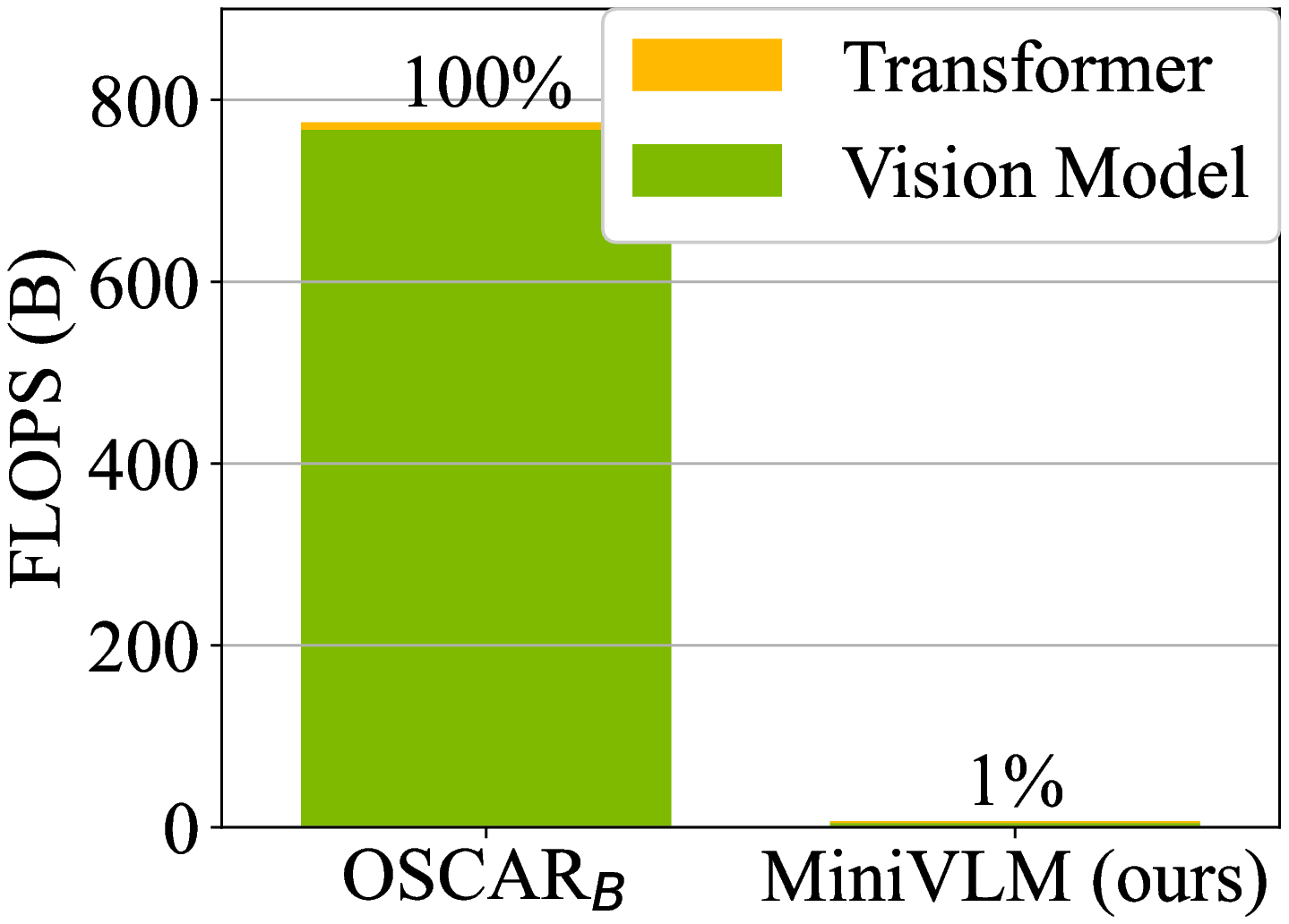} \\
	\multicolumn{2}{c}{
	\includegraphics[width=0.96\linewidth]{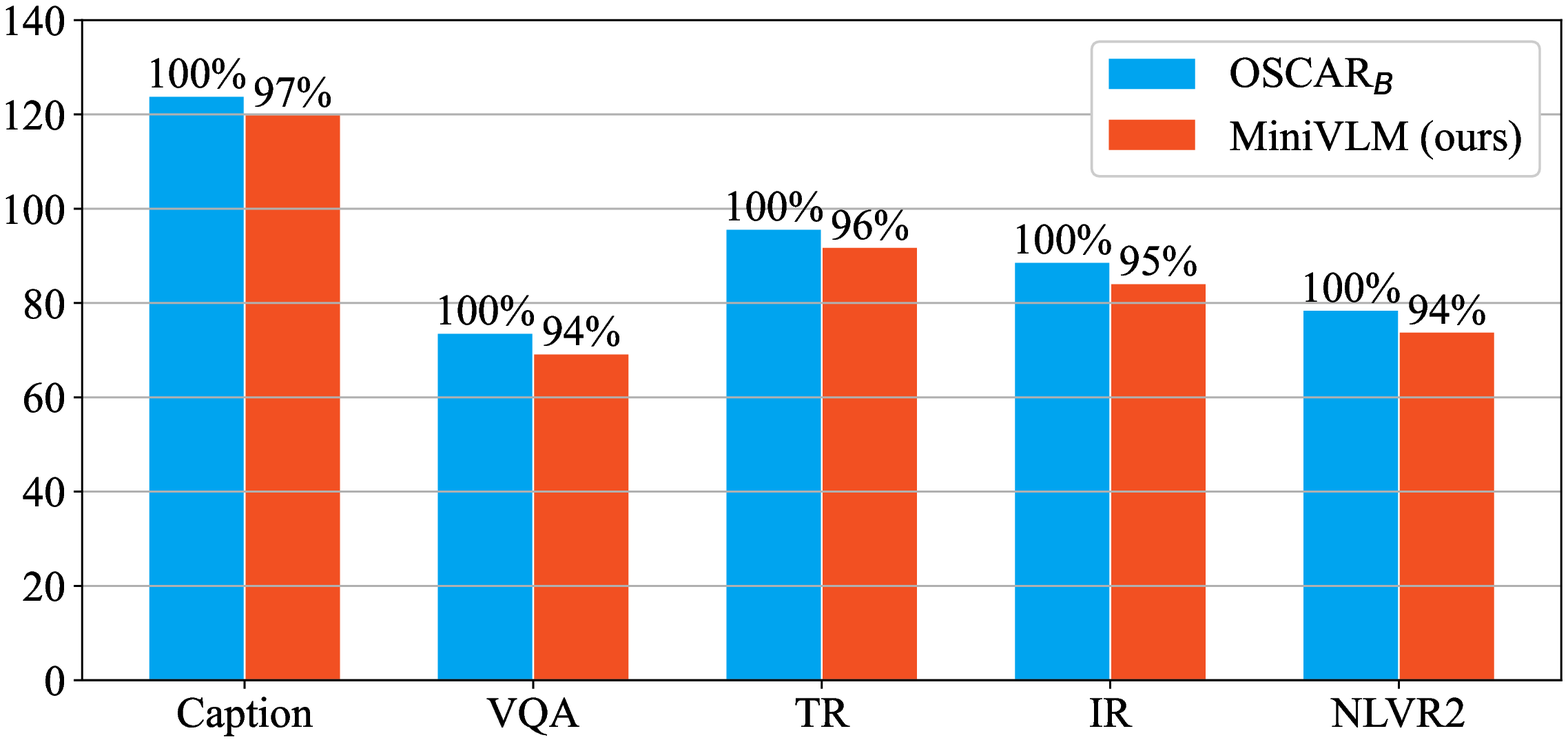}
	}
	\end{tabular}
   \caption{MiniVLM retains $94-97\%$ of the accuracy on multiple tasks 
   with $27\%$ parameters and $1\%$ FLOPS compared to state-of-the-art model
   OSCAR$_\text{B}$~\cite{Li0LZHZWH0WCG20}.
   Details can be found in Sec.~\ref{sec:downstream}.
   }
   \label{fig:trade_off}
\end{figure}

With the success of BERT~\cite{DevlinCLT19} and recent advances~\cite{ZhouPZHCG20,LuBPL19,SuZCLLWD20,TanB19,abs-1908-03557,LiDFGJ20,Li0LZHZWH0WCG20,abs-2009-13682,ChenLYK0G0020,abs-2004-00849,abs-1711-07264} 
in vision-language pre-training (VLP), models pre-trained on large-scale image-text data have made substantial improvement on various benchmarks for a wide range of vision-language (VL) tasks, such as image captioning, visual question answering and image-text retrieval.
The models used in most VLP works contain two modules: the vision module based on convolutional neural networks trained on ImageNet~\cite{RussakovskyDSKS15} and/or Visual Genome (VG)~\cite{KrishnaZGJHKCKL16} to extract visual features from the image; and the feature fusion module based on the multi-modal transformer model to process both the visual features and the token embeddings of the text input.
The VL models are firstly pre-trained to learn cross-modal representations, and then fine-tuned on task-specific data.
In recent VLP research, both of the two modules leverage large-scale deep neural networks, which can take up to hundreds of millions of parameters, to achieve the state-of-the-art performance. However, due to the large sizes and high computation cost, it could be impractical for real-world applications to exploit the power of large models under a constrained training and/or inference budget.
In fact, building a lightweight VL model, which is desired when operating on resource-limited devices, is of great practical value but is less explored in the literature.

While larger models have been demonstrated to achieve higher performance in extensive studies, it is challenging to compress the model to smaller sizes without tremendous performance drop.
In order to retain as much performance as possible, we firstly optimize the network architecture to balance accuracy and speed. Moreover, we improve the small-model pre-training by leveraging large models and large-scale dataset.

For the architecture of VL models,
popularized as ``bottom-up top-down'' (BUTD) attention~\cite{00010BT0GZ18}, most existing works~\cite{ZhouPZHCG20,LuBPL19,SuZCLLWD20,TanB19,abs-1908-03557,LiDFGJ20,Li0LZHZWH0WCG20,abs-2009-13682,ChenLYK0G0020} use the ResNet-101 Faster R-CNN~\cite{RenHGS15} model trained on the VG dataset as the visual feature extractor, which has been well validated by state-of-the-art results on various benchmarks.
However, the detector suffers from a heavy model size and high latency, and consequently cannot be deployed to resource-limited applications. A few recent works~\cite{abs-2004-00849,abs-1711-07264} revisit the usage of grid features from the convolutional layer to skip the region-related computation in Faster R-CNN. Nevertheless, it is still an open problem to select over the overwhelming number of grid features, as dumping the whole feature map to transformer could be prohibitively expensive in computation.
For the transformer module, BERT is widely used as the de facto standard. Recent work in Natural Language Processing (NLP) has explored to maintain high performance with compact structures based on BERT.
However, the compact structure is less investigated in VLP.

In this paper, we propose a smaller and faster VL model, named MiniVLM, to reach similar performance as its larger counterpart with a much smaller size, resulting in faster inference speed.
For the vision module in MiniVLM, we design a \textbf{T}wo-stage \textbf{E}fficient feature \textbf{E}xtractor (TEE) to drastically reduce the computation cost for extracting visual features, which is a dominating part of the inference cost on certain tasks.
While refining each part of the detection model, we greatly simplify the region-related modules in TEE. The underlying implication is that the VL tasks require rich visual representations rather than precise box locations as in the object detection task.
Experimental results show that our TEE can extract visual features of similar quality at a much faster speed. In particular, our TEE-0, using a similar backbone as EfficientDet-D0~\cite{TanPL20}, is $3.7 \times$ smaller and $99 \times$ faster than the widely used R101 Faster R-CNN from BUTD,
while retaining competitive accuracy in detection on VG, and up to $97\%$ of the accuracy on downstream tasks.
For the transformer model, we choose the MiniLM\cite{abs-2002-10957} structure after empirically evaluating the performance of several structures, including BERT~\cite{DevlinCLT19} and its compact variants~\cite{abs-1910-01108,SunYSLYZ20,abs-1909-10351,abs-2002-10957}.

In addition to the model architecture optimization, we leverage high-accuracy large models and large-scale data, either labeled or unlabeled, to further boost the performance of the small pre-trained model.
To improve the accuracy of TEE, we pre-train it on 
large-scale classification and detection dataset
before fine-tuning on VG.
During the VL pre-training, we apply data distillation~\cite{RadosavovicDGGH18,XieLHL20,abs-1905-00546} to add $7M$ Open Images~\cite{OpenImages} which are pseudo-labeled by the state-of-the-art “teacher” captioning model. 
We also use the high-quality tags from a large tagging model in pre-training to improve visual-text alignment.
The large tagging model is not used in fine-tuning or inference, and therefore has no impact on the runtime speed. With the above ingredients, our MiniVLM, composed of TEE-0 and MiniLM~\cite{abs-2002-10957}, reduces the end-to-end FLOPs to $1\%$ with $27\%$ parameters, and retains $94-97\%$ accuracy compared to large state-of-the-art models on multiple VL tasks.

In summary, we make the following contributions.
\begin{itemize}
  \setlength\itemsep{0em}
   \item We propose a VL model MiniVLM, which can be fine-tuned with good performance on multiple downstream tasks, while being smaller and faster for practical application. 
  \item We design a Two-stage Efficient feature Extractor (TEE) to extract image region features for VL tasks, which generates features of good quality at a much faster speed.
  \item We demonstrate the benefits of using large models as well as large-scale data in the small VL model pre-training stage to improve the downstream tasks. 
\end{itemize}

\begin{figure*}[t]
\begin{center}
  \includegraphics[trim=80 35 120 250, clip,width=1\textwidth]{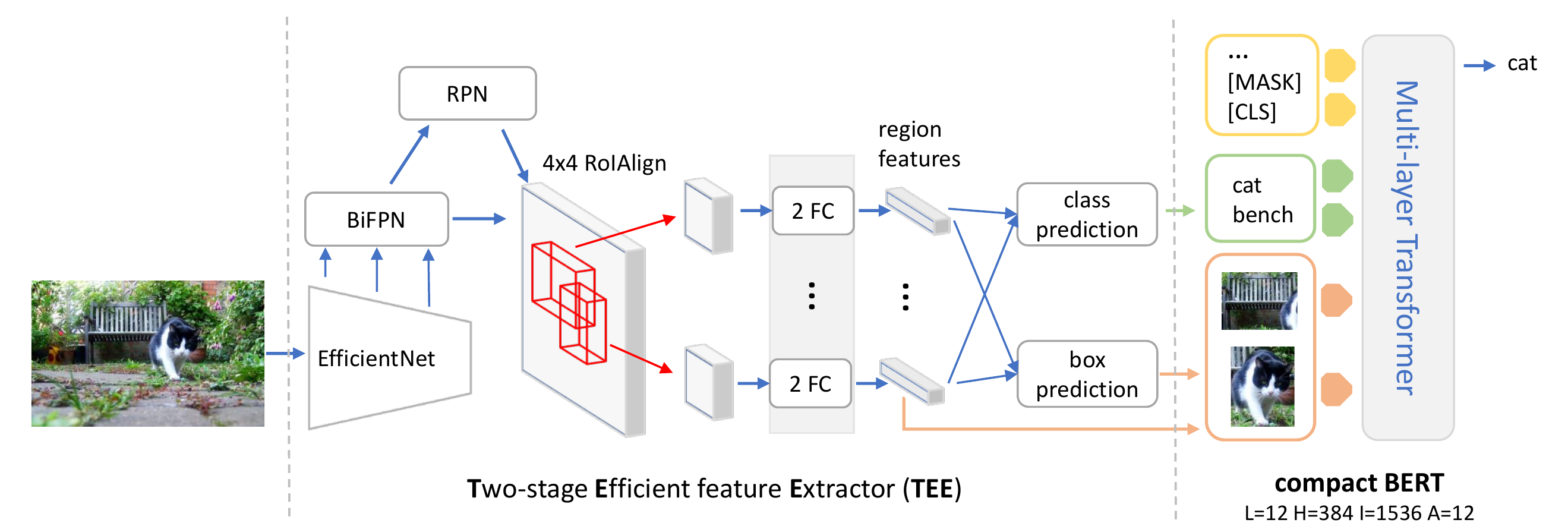}
\end{center}
\caption{
The proposed MiniVLM architecture, consisting of the Two-stage Efficient feature Extractor (TEE) and compact BERT feature fusion module. During inference, the vision module detects objects and extracts region features, which are fed to the transformer model. The text input of the transformer also depends on the downstream task. For image captioning, [CLS] is given as the first token, then the model predicts the next token in an auto-regressive manner to generate a caption sentence.
} 
\label{fig:arch}
\end{figure*}

\section{Related work}

\noindent\textbf{Vision-Language Pre-training.}
Remarkable progress~\cite{LuBPL19,SuZCLLWD20,abs-2004-00849,lin2020interbert,abs-1908-03557,Li0LZHZWH0WCG20,abs-2009-13682,ChenLYK0G0020} has been made
recently on vision-language tasks through network pre-training 
on massive data with image-text pairs.
A popular framework used in most VLP work is to view the extracted visual features as visual `tokens' and feed them 
together with text tokens into the BERT~\cite{DevlinCLT19,VaswaniSPUJGKP17} model
for joint representation learning.

The visual feature is generally extracted with an off-the-shelf vision model, and the main focus is on the multi-modal fusion based on BERT model.
With the multiple modalities, the fusion can
be categorized as early fusion, late fusion and full fusion. 
Early fusion is to first process the two modalities 
together and then process each 
separately to enhance the single-modality task, 
\eg InterBERT~\cite{lin2020interbert}.
Late fusion is to first process each modality separately and then to fuse them 
together, \eg in ViLBERT~\cite{LuBPL19},
LXMERT~\cite{TanB19},
ERNIE-ViL~\cite{abs-2006-16934}.
Full fusion means to process the two modalities' features 
together with the BERT model from the very beginning to the final representation, \eg
OSCAR~\cite{Li0LZHZWH0WCG20},
Unicoder-VL~\cite{LiDFGJ20},
VL-BERT~\cite{SuZCLLWD20},
UNITER~\cite{ChenLYK0G0020},
VIVO~\cite{abs-2009-13682}.
The pre-training tasks typically include the masked language modeling, image-text pairing loss, and masked region modeling.

\noindent\textbf{Visual Feature Extractor.}
Visual feature extraction is one of the key modules in vision-language (VL) tasks.
As in the bottom-up top-down approach~\cite{00010BT0GZ18}, region features based on 
Faster R-CNN~\cite{RenHGS15} have shown strong accuracy 
and been 
widely used in VL tasks~\cite{ZhouPZHCG20,LuBPL19,SuZCLLWD20,TanB19,abs-1908-03557,LiDFGJ20,Li0LZHZWH0WCG20,abs-2009-13682}.
The extractor is trained on ImageNet~\cite{RussakovskyDSKS15} and 
Visual Genome~\cite{KrishnaZGJHKCKL16} datasets with two training tasks: one is to predict the object category and 
the other is to predict the attribute information. 

An alternative approach is the grid feature, which is revisited in~\cite{jiang2020defense,abs-2004-00849}
and demonstrated encouraging performance.
In~\cite{jiang2020defense}, the grid feature extractor is constructed by 
casting 
the Faster R-CNN model into a fully-convolutional network and remove the region-related
operations (\eg non-suppressed compression) to reduce the time cost. 
In~\cite{abs-2004-00849}, the convolutional network is trained together with modality fusion 
network without the detection data.

One advantage of using region features is that it is easy to 
select the top-$K$ salient regions
as each region is associated with a confidence score.
Typically, the number of region features is $50$ while 
the number of grid features ranges from $300$ to $600$ 
as in~\cite{jiang2020defense}.
With more features, the cost of the multi-modal fusion part can be increased significantly.
Thus, in this paper, we stick to the region features for our compact model.

\noindent\textbf{Object Detection.}
Region feature is built on the object detector, and the detector
can be two-stage~\cite{RenHGS15,HeGDG20,abs-1711-07264,0008CYLWL020} or one-stage~\cite{RedmonDGF16,TanPL20,LiuAESRFB16,LinGGHD20,GhiasiLL19,ZhangCYLL20,WangACKP019,abs-2006-09214}.
The two-stage detector generates
bounding box candidates with a region proposal network (RPN) and 
extracts the region features with RoIPool~\cite{RenHGS15} or RoIAlign~\cite{HeGDG20}.
The feature is further processed with a classification head and a bounding box regression head.
In contrast, the one-stage detector removes the RPN, and 
predicts the bounding box results based on the convolutional neural network directly.

Due to the removal of RPN and region feature extraction,
fast object detector are mostly based on one-stage detectors, \eg~\cite{RedmonDGF16,TanPL20,Wang0AL18,MehtaRSH19,abs-1904-08900}.
However, it remains open on how to effectively extract region features directly from one-stage detectors
for VL models.
Thus, we use a two-stage architecture but design a lightweight backbone and detection head for the compact VL model.

\noindent\textbf{Compact BERT.}
BERT$_\text{BASE}$ or BERT$_\text{LARGE}$ has been commonly 
used in the existing VL works.
To reduce the cost, one can simply reduce the network dimensions, \eg the number of layers, the hidden size, as in TinyBERT~\cite{abs-1909-10351} and  MiniLM~\cite{abs-2002-10957}.
MobileBERT~\cite{SunYSLYZ20} constructs the network with the bottleneck 
design~\cite{HeZRS16} to reduce the cost.
ALBERT~\cite{abs-1909-11942} focuses on the reduction of the parameter size. 
In our compact solution,  we choose 
MiniLM~\cite{abs-2002-10957} as our multi-modal fusion module
after comparing different approaches in VL tasks.

\noindent\textbf{Data Distillation.}
Data distillation (and self-training) is a simple yet effective approach
to leverage massive raw images
with pseudo labels generated from a strong pre-trained model.
The effectiveness has been well demonstrated, \eg in 
image classifications~\cite{abs-1905-00546,abs-1909-11942}
and object detection~\cite{RadosavovicDGGH18}.
Here we apply data distillation to the VL model. 
One potential improvement is to apply the model distillation or knowledge distillation~\cite{HintonVD15} on both the vision module and the transformer fusion module, which we leave as future work. 

\section{MiniVLM}

In this section, we describe how we design a smaller and faster VL model, MiniVLM, and improve the accuracy 
for small VL model.
An overview of our model is shown in Fig.~\ref{fig:arch}. It consists of a detector-based feature extractor and a transformer-based feature fusion module.
For various downstream tasks, we alter the transformer prediction head with minimal changes, which we defer to Sec.~\ref{sec:downstream}.


\subsection{Model architecture}\label{sec:tee}
\Paragraph{Two-stage Efficient feature Extractor.} 
While the R101 Faster R-CNN detector from~\cite{00010BT0GZ18} has been widely used to extract region features, the computational cost is largely overlooked, which can take a majority of the total inference time for some VL tasks. 
Although region feature extraction is part of an objection detection model, the requirement for VL tasks is not the same as for objection detection. For VL tasks, the transformer
is used to
reason the relationship between
visual and language semantics, and what is needed from the feature extractor is 
rich visual representations. For example,
the bounding box locations do not have to be highly accurate, and
the recall of the bounding boxes is more important 
to cover more visual information from the image.
These characteristics allow us to design a feature extractor that is much more efficient while not causing significant accuracy degradation for the downstream tasks. Fig.~\ref{fig:arch} shows the design of our feature extractor called Two-stage Efficient feature Extractor (TEE). 

First, we replace the backbone with EfficientNet~\cite{TanL19} and add BiFPN~\cite{TanPL20} to generate multi-scale features.
Both components consists of depthwise and pointwise convolutional layers, which reduce the model size and computation significantly compared with the standard convolutional layers.
The BiFPN receives as input $4$ layers with $\text{stride} = 4, 8, 16, 32$ from EfficientNet, and outputs $5$ features with an extra feature map of $\text{stride}=64$ by downsampling.
Both EfficientNet and BiFPN are building blocks of the one-stage detector EfficientDet~\cite{TanPL20}, while we make the change to use feature maps starting from $\text{stride}=4$ instead of $8$ to incorporate information from higher resolution feature maps for the feature extraction.

While region proposal network (RPN)~\cite{RenHGS15} is used following the design of two-stage detectors, the box prediction modules are greatly simplified. Our RPN contains only $2$ convolutional layers with kernal size as $1$: one for bounding box regression and the other for objectness prediction.
After non-maximal suppression (NMS) we select the feature map for each box proposal with heuristics from~\cite{LinDGHHB17}, and apply RoIAlign~\cite{HeGDG20} operation, followed by $2$ linear layers to extract the region features. The resolution of RoIAlign is reduced to $4\times 4$
rather than $14\times 14$ in~\cite{RenHGS15,00010BT0GZ18} 
or $7\times 7 $ in~\cite{LinDGHHB17}.
The feature's dimension is also reduced from $2048$~\cite{RenHGS15,00010BT0GZ18} to $1024$.
In~\cite{00010BT0GZ18}, NMS is applied for each class which can be up to $1600$ times on Visual Genome~\cite{KrishnaZGJHKCKL16}. To reduce the cost, one can apply sophisticated approaches, \eg~\cite{abs-2005-11426} or~\cite{sangBS17,peize2020sparse,CarionMSUKZ20} to remove NMS.
For simplicity, we apply NMS once in a class-agnostic manner to save the computation.

Similar to EfficientDet, we scale up the input image size, network depth and width to get stronger feature extractors.
For varying EfficientDet-DX (X = 0, 1, $\cdots$), the corresponding TEE is denoted as TEE-X.
Without confusion, we also use TEE to refer to TEE-0 as our extractor for MiniVLM. 


During the inference, given an image $\bf{I}$, the vision module outputs a bag of region features ${\bf{R}}$ with corresponding bounding boxes ${\bf{B}}$ and object tags ${\bf{C}}$, which are fed to the transformer model along with text tokens.
It is noted that no extra tagging model is employed. We re-use the feature extractor as the tagging model and treat the region class names as the tags. 

\Paragraph{Multi-modal Transformer.} 
With the extracted features, a transformer-based feature fusion module is applied.
To strike a good balance between speed and accuracy, 
we search the compact structures based on BERT by varying some parameters, \eg, the number of layers.
Based on experimental results, we
choose the same structure as MiniLM~\cite{abs-2002-10957}, \ie, $12$ layers with
hidden size reduced to $384$ and feed-forward intermediate size reduced to $1536$.
We follow~\cite{Li0LZHZWH0WCG20} to train the transformer model. The input 
consists of visual features
formed by the concatenation of $\mathbf{R}$
and bounding box encoding (normalized $4$ corner coordinates and the heigh/width of the box),
tokenized object tag names ${\bf{C}}$, and tokenized sentences ${\bf{S}}$.
The content of ${\bf{S}}$ can vary depending on the downstream task, \eg, the question sentence for VQA, a single \texttt{[CLS]} token to indicate the start of sentence for image captioning.

\subsection{Pre-training}\label{sec:bert}
To train a VL model, the vision module is first trained on classification or detection dataset to learn diverse visual representations. Given the visual features, the transformer module is then pre-trained on massive image-text pairs to learn cross-modal representations. Finally, the model is fine-tuned on specific downstream tasks.
To compensate the performance drop brought 
about 
by the small model size, we apply several techniques in training.

As visual features are critical in VL tasks, we improve visual feature by 
pre-training TEE on large-scale classification and object detection dataset, \eg, Objects365~\cite{0005LZPYZLS19}, before fine-tuning on the Visual Genome dataset,
which shows the performance gain for various downstream VL tasks

By pre-training the transformer model on large-scale image-text data,
our model inherits the advantage of VL pre-training. Moreover, 
we leverage large models in two ways to further exploit the potential for pre-training with compact VL models.
First, 
we apply a state-of-the-art captioning model to describe $7M$ images from Open Images with pseudo captions. In this way, the small model learns to mimic the behavior of the large model through much more data, which can be further expanded with internet-scale unlabeled data. Second, we use a large tagging model to generate high-quality tags, and also add ground truth tags if available. Although the tags in pre-training are from different sources,
the tags in fine-tuning are generated by the same vision model used to extract features
to remove the dependency on the large model at inference time. 
The experimental results in Sec.~\ref{sec:exp:analysis} shows that the better quality of tags helps with cross-modal representation learning.

Other than the changes about the sources of 
object tags and the associated sentences,
we use the same pre-training tasks as described in~\cite{Li0LZHZWH0WCG20}, including masked language modeling (MLM) and image-text (contrastive) matching (ITM).

\begin{table}[]
    \centering
    \begin{tabular}{@{}ll@{~}c@{~}c@{~}c@{}}
    \toprule
    \multicolumn{2}{c}{Model}                               &  Params(M)    & FLOPS(B)    & $\text{mAP}_{0.5}$    \\
    \midrule
    \multirow{2}{*}{Grid~\cite{jiang2020defense}} & R50     & $23.5$         & $37.8$       & -    \\
                                                  & X101    & $86.9$         & $161.2$      & - \\
    \midrule
    \multirow{2}{*}{Region} & R101-F~\cite{00010BT0GZ18}    & $63.8$         & $767.0$       & $10.2$\tablefootnote{This number is from https://github.com/peteanderson80/bottom-up-attention.}    \\
    & TEE (ours)                                           & $7.5$          & $4.4$         & $9.9$  \\
    \bottomrule
    \end{tabular}
    \caption{
    Comparison of different vision modules on number of parameters, FLOPS, and detection accuracy on VG.
    }
    \label{tab:detector}
\end{table}

\section{Experiment}
\subsection{Implementation details}
\noindent\textbf{TEE.}
We first pre-train the backbone on ImageNet~\cite{RussakovskyDSKS15} classification dataset, then pre-train the whole detection model on Objects365~\cite{0005LZPYZLS19}, and lastly fine-tune it on Visual Genome~\cite{KrishnaZGJHKCKL16}.
On ImageNet, the backbone is trained from scratch for $400$ epochs.
Stochastic gradient descent (SGD) is used to optimize the model with the batch size of $1024$. The learning rate is $0.4$ and decays with a cosine scheduler~\cite{LoshchilovH17}.
Afterwards, the detection model
is initialized with this ImageNet-pretrained backbone and trained on Objects365 for $100$ epochs. The learning rate is $0.4$ and batch size is $256$ with SGD.
Lastly, we fine-tune the model on Visual Genome for $200$ epochs, with learning rate $0.2$ and batch size $512$. 
Following~\cite{00010BT0GZ18},
an additional head is added to train with attribute classes.

\noindent\textbf{Vision-Language Pre-training.} 
We combine existing V+L datasets, including MS COCO~\cite{LinMBHPRDZ14}, Conceptual Captions (CC)~\cite{SoricutDSG18}, 
SBU captions~\cite{OrdonezKB11}, Flicker30k~\cite{YoungLHH14}, GQA~\cite{HudsonM19},
VQA~\cite{GoyalKSBP16} and VG-QA~\cite{KrishnaZGJHKCKL16}.

To explore data distillation for the compact VL model pre-training, we incorporate the
Open Images V6~\cite{OpenImages} as extra images and generate pseudo captions using the state-of-the-art image captioning model fine-tuned from OSCAR~\cite{Li0LZHZWH0WCG20}.
The human verified positive tags are combined with object class predictions from TEE-3, and together serve as the tag input in our VL pre-training. This dataset is referred to as OI-Caps-7M. 



During VL pre-training, 
the batch size is $2048$, and the initial learning rate is $4\times 10^{-4}$ with linear decay. 
The model is updated with AdamW~\cite{LoshchilovH19} optimizer for 100 epochs.


\begin{table}[t!]
    \centering
    \begin{tabular}{@{}l@{~}c@{~}c@{~}c@{~}c@{~}c@{~}c@{}}
    \toprule
       Model                                 & Config           & Params(M)  & FLOPS(B) \\
       \midrule 
       BERT$_\text{BASE}$ \cite{DevlinCLT19} & $12/768/3072$    & $134.3$    & $8.2$ \\
       \midrule
       BERT$_8$                              & $8/768/3072$     & $106.0$    & $5.8$ \\
       TinyBERT$_6$~\cite{abs-1909-10351}    & $6/768/3072$     & $91.8$     & $4.6$ \\
       BERT$_4$                              &  $4/768/3072$    & $77.6$     & $3.3$ \\
       MiniLM~\cite{abs-2002-10957}          & $12/384/1536$    & $45.7$     & $2.3$ \\
       TinyBERT$_4$~\cite{abs-1909-10351}    & $4/312/1200$     & $24.3$     & $0.8$ \\
       \bottomrule
    \end{tabular}
    \caption{Computational cost for different transformer structures.
    Config: the number of layers, the embedding dimension, and intermediate size. 
    FLOPs: measured in one forward pass with $50$ image regions and $35$ text tokens.}
    \label{tab:bert_cost}
\end{table}


\subsection{Smaller and faster}\label{sec:exp:time}
As shown in Fig.~\ref{fig:trade_off}, 
our MiniVLM reduces the number of parameters to $27\%$
and FLOPS to $1\%$ in total compared to the model in~\cite{Li0LZHZWH0WCG20}. 
The following details the compression for both the vision module and the transformer module.

\begin{table*}[t!]
    \centering
    \begin{tabular}{lccc@{~}cc@{~}c@{~}c@{~}cc}
    \toprule
    \multirow{2}{*}{Method}                     &  \multicolumn{2}{c}{Network}      & \multicolumn{2}{c}{Cost}   & \multicolumn{4}{c}{Evaluation on Captioning} & VQA \\
                                                \cmidrule(lr){2-3}              \cmidrule(lr){4-5}                  \cmidrule(lr){6-9} \cmidrule(lr){10-10}
                                                & Vision  & Transformer             &Params(M)$\downarrow$ &FLOPs(B)$\downarrow$&B@4$\uparrow$&M$\uparrow$&C$\uparrow$&S$\uparrow$ & Acc. $\uparrow$\\
    \midrule
    OSCAR$_{\text{B}}$~\cite{Li0LZHZWH0WCG20}   & R101-F  & BERT$_\text{BASE}$      & $198.1$   &  $775.2$        & $36.5$   & $30.3$    & $123.7$  & $23.1$ & $73.16$\\
    \midrule
     BERT $\rightarrow$ MiniLM                  & R101-F  & MiniLM                  & $109.5$   &  $769.3$        & $35.0$   & $28.5$    & $118.8$  & $21.7$ & $68.69$\\
    R101-F $\rightarrow$ TEE-0                  & TEE-0   & BERT$_\text{BASE}$      & $141.8$   & $12.6$         & $34.6$   & $28.4$    & $118.7$  & $21.6$  &  $69.31$ \\
    \midrule
    baseline                                    & TEE-0   & MiniLM                  & $53.2$    & $6.7$          & $34.0$   & $27.8$    & $115.0$  & $21.2$  &  $68.14$   \\
    \midrule
    +O                                          & TEE-0   & MiniLM                  & $53.2$    & $6.7$          & $34.3$   & $28.1$    & $116.7$  & $21.3$ & $68.54$ \\
    +O + H                                      & TEE-0   & MiniLM                  & $53.2$    & $6.7$          & $34.7$   & $28.3$    & $117.7$  & $21.4$ &  $68.85$ \\
    +O + H + DD (MiniVLM)                       & TEE-0   & MiniLM                  & $53.2$    & $6.7$          & $35.6$   & $28.6$    & $119.8$  & $21.6$ &  $69.09$ \\
    \bottomrule
    \end{tabular}
    \caption{Retaining high accuracy from OSCAR$_\text{B}$ to MiniVLM on the image captioning (Karpathy split) 
    and VQA 2.0 task (\texttt{test-dev}). 
    O: using Objects365 to pre-train TEE before fine-tuning on Visual Genome. H: using high-quality tags during vision-language pre-training. DD: with data distillation by adding OI-Caps-7M to the pre-training corpus.
    For the metrics, $\uparrow$ indicates higher is better, $\downarrow$ indicates lower is better. Captioning results are evaluated with BLEU-4 (B@4), METEOR (M), CIDEr (C) and SPICE (S).
    }
    \label{tab:ablation}
\end{table*}

\noindent\textbf{TEE.} 
We compare our region feature extractor TEE
with 
the widely used ResNet-101 Faster R-CNN model (R101-F) from ~\cite{00010BT0GZ18},
as well as the grid feature extractor based on ResNet-50 and ResNeXt-101 from ~\cite{jiang2020defense}.
Table~\ref{tab:detector} shows the size and computation cost for each model at inference time.
Compared to R101-F, our TEE reduces the number of parameters to $7.5/63.8=11.8\%$,
and FLOPS to $4.4/767.0=1\%$. 
Table~\ref{tab:param} and Table~\ref{tab:flops} show the breakdown on each component of the parameters and FLOPs, respectively. For R101-F, the major cost resides in the box head, which consists of $3$ residual blocks and the number of output channels in each block is $2048$.
In the backbone, the largest number of channels is $1024$, and thus the box head is even more expensive than the backbone. 
In comparison, the box head of our TEE only contains $2$ linear layers, which significantly reduces the cost. 
Our model also uses fewer parameters and FLOPS than grid feature extractors.
The reason is the use of the depthwise and pointwise convolutional layers in backbone
and the lightweight region feature extraction head. 

\begin{table}[]
    \centering
    \begin{tabular}{ccccc}
    \toprule
     Extractor                    &  Backbone     & RPN      & Box head      & Total \\
                              \cmidrule(){1-1}\cmidrule(lr){2-4} \cmidrule(){5-5}
    R101-F~\cite{00010BT0GZ18}  & $27.6$          & $4.7$      & $31.4$          & $63.8$ \\
    TEE-0                       & $3.8$           & $10^{-3}$   & $3.7$           & $7.5$ \\
    \bottomrule
    \end{tabular}
    \caption{Number of parameters (in million) in each component of the region feature extractors.}
    \label{tab:param}
\end{table}

\begin{table}[]
    \centering
    \begin{tabular}{ccccc}
    \toprule
    Extractor                   &  Backbone     & RPN      & Box head       & Total \\
                              \cmidrule(){1-1}\cmidrule(lr){2-4} \cmidrule(){5-5}
    R101-F~\cite{00010BT0GZ18}  &  $67.1$       &  $9.1$   & $690.8$        & $767.0$ \\
    TEE-0                       & $3.3$         & $0.03$   & $1.1$          & $4.4$ \\
    \bottomrule
    \end{tabular}
    \caption{FLOPs (in billion) in each component of the region feature extractors.}
    \label{tab:flops}
\end{table}

On a CPU workstation\footnote{Intel(R) Xeon(R) CPU E5-2620 v4 @2.10GHz} with 4 threads, with models implemented in PyTorch\footnote{https://github.com/pytorch/pytorch} and processing one image at a time,
Grid R50 takes $699.8\pm 110.1$ ms, R101-F takes $12.3\pm 3.2$ seconds, while our TEE takes only $393.9 \pm 43.8$ ms,
which is $3.2\%$ of R101-F. 
Note that inference speed highly depends on hardware and implementation, so we mainly
report FLOPS for fair comparison.

\noindent\textbf{Compact BERT.}
Table~\ref{tab:bert_cost} shows the comparison of different 
transformer model structures, including
TinyBERT~\cite{abs-1909-10351} with $4$ and $6$ layers, 
MiniLM~\cite{abs-2002-10957}, 
BERT$_\text{BASE}$ with the number of layers cut to $4$ (BERT$_4$) and $8$ (BERT$_8$), and BERT$_\text{BASE}$.
The whole transformer model, including embedding and decoding layers, is counted.
By choosing MiniLM, which will be explained later with Fig.~\ref{fig:bert}, 
the number of parameters is reduced to $45.7/134.3=34.0\%$, and FLOPS
to $2.3/8.2=28.0\%$.
As to the inference time evaluated on the image captioning task, 
BERT$_\text{BASE}$ takes $712.6\pm 176.1$ ms to process one image,
while MiniLM takes $346.7 \pm 49.8$ ms, reducing to $346.7/712.6=48.7\%$.


\subsection{Retaining high accuracy}
Table~\ref{tab:ablation} shows the ablation study on improving the pre-training for our small model.
The pre-trained models are fine-tuned and evaluated on the image captioning and VQA task, which will be detailed in Sec.~\ref{sec:downstream}.
The cost is measured end-to-end including both vision and transformer modules.
Starting from the OSCAR$_\text{B}$~\cite{Li0LZHZWH0WCG20} model,
which consists of R101-F and BERT$_\text{BASE}$, 
if we replace the transformer module with MiniLM,
the CIDEr score drops $4.9$ points.
If solely replacing the vision module with TEE-0,
the CIDEr score drops similarly $5.0$ points.
This also indicates that our feature extractor TEE-0 can achieve $118.7/123.7=96\%$ of the accuracy compared to R101-F on this task without any additional techniques.
Then, we replace both modules to 
TEE-0 and MiniLM, respectively, where the CIDEr score is decreased by $8.7$.
This is the baseline performance of our compact VL model.

Next, we apply approaches in Sec.~\ref{sec:bert} and show the improvement for pre-training with small-model.
First, we use the Objects365 dataset to 
pre-train TEE before fine-tuning it on VG, which 
improves the CIDEr score by $1.7$,
indicating that better visual features contribute to better performance on VL tasks.
Secondly, we use high-quality tags, generated from the stronger vision model TEE-3, during vision-language pre-training, and further improve the score by $1.0$.
The intuition is based on~\cite{Li0LZHZWH0WCG20} that the tag information in pre-training helps with visual-text alignment.
Lastly, we add the OI-Caps-7M dataset, and observe the gain of $2.1$ in CIDEr.
In total, the CIDEr score is improved by $4.8$, resulting in a much smaller gap with the large pre-trained VL model.
Similar trend can be observed for the VQA task as shown in the last column of Table~\ref{tab:ablation}.


\subsection{Results on downstream VL tasks}\label{sec:downstream}

\noindent\textbf{Image Captioning.} The task is to describe an image with a natural language sentence. Following~\cite{ZhouPZHCG20,Li0LZHZWH0WCG20}, we fine-tune the model with region features, captioning tokens and object tags. Captioning tokens are randomly replaced by the token of \texttt{[MASK]} with $15\%$ chance and predicted by the corresponding representation, the attention of which is only on region features, tags and preceding caption tokens. The training task is either the cross entropy loss or the loss optimized for the CIDEr~\cite{VedantamZP15} score, and we report results on both tasks.  
During inference, the \texttt{[MASK]} is appended recursively with the generated tokens to predict the next token one by one. Considering the inference speed, we reduce the beam search size to 1 instead of 5 as in~\cite{Li0LZHZWH0WCG20}.
The accuracy is evaluated with BLEU@4~\cite{PapineniRWZ02}, METEOR~\cite{DenkowskiL14}, CIDEr~\cite{VedantamZP15}, and SPICE~\cite{AndersonFJG16}. The dataset is COCO~\cite{LinMBHPRDZ14} with Karpathy split~\cite{KarpathyL15}.

\begin{table*}[t]
	\begin{tabular}{c@{~~}c}
	\parbox{0.55\linewidth}{
		\centering

	\begin{tabular}{l@{~~}c@{~~}c@{~~}c@{~~}c@{~~}c@{~~}c@{~~}c@{~~}c@{}}
		\toprule
		\multirow{2}{*}{Method}                                 & \multicolumn{4}{c}{CE Optimization} & \multicolumn{4}{c}{CIDEr optimization}    \\ 
		\cmidrule(lr){2-5}                               \cmidrule(lr){6-9}                         
		& B@4    & M       & C        & S                & B@4    & M      & C        & S            \\
		\midrule
		BUTD~\cite{00010BT0GZ18}                                & $36.2$ & $27.0$  & $113.5$  & $20.3$           & $36.3$ & $27.7$ & $120.1$  & $21.4$       \\
		Grid~\cite{jiang2020defense}                            & $36.4$ & $27.4$  & $113.8$  & $20.7$           & -      & -      &-         & -            \\
		AoANet~\cite{huang2019attention}                        & $37.2$ & $28.4$  & $119.8$  & $21.3$           & $38.9$ & $29.2$ & $129.8$  & $22.4$       \\
		OSCAR$_{\text{B}}$~\cite{Li0LZHZWH0WCG20}               & $36.5$ & $30.3$  & $ 123.7$ & $23.1$           & $40.5$ & $29.7$ & $137.6$  & $22.8$       \\
		\midrule  
		MiniVLM (Ours)                                          & $35.6$ & $28.6$  & $119.8$  & $21.6$           & $39.2$ & $29.7$ & $131.7$  & $23.5$       \\
		\bottomrule
	\end{tabular}
	\captionsetup{skip=0pt}
	\captionsetup{width=1.\linewidth}
	\caption{Image captioning evaluation results (single model) on COCO `Karpathy'~\cite{KarpathyL15} test split. (B@4: BLEU@4, M: METEOR, C: CIDEr, S: SPICE.)}
	\label{tab:coco_caption}
	}
& 
	\parbox{.4\linewidth}{

	\begin{tabular}{l@{~~}c@{~}c@{~~}c@{~}c}
		\toprule
		\multirow{2}{*}{Method}                                 & \multicolumn{2}{c}{VQA}   & \multicolumn{2}{c}{NLVR2}\\ 
		\cmidrule(lr){2-3}        \cmidrule(lr){4-5}
		& test-std  & test-dev      & Test-P  & Dev \\
		\midrule
		BUTD~\cite{00010BT0GZ18}                                & $70.34$   & -             & -       &   -     \\
		Grid~\cite{jiang2020defense}                            &   -       & $72.59$       & -       &   -     \\
		Pixel. (R50)~\cite{abs-2004-00849}                   & $71.42$   & $71.35$       & $71.7$  & $72.4$  \\
		Pixel. (X152)~\cite{abs-2004-00849}                  & $74.55$   & $74.45$       & $76.5$  & $77.2$  \\
        VisualBERT~\cite{abs-1908-03557}                        & $71.00$   & $70.80$        & $67.4$  & $67.0$  \\
        OSCAR$_{\text{B}}$~\cite{Li0LZHZWH0WCG20}               & $73.44$   & $73.16$       & $78.07$ & $78.36$ \\
		\midrule  
		MiniVLM (Ours)                                          & $69.44$   & $69.09$       & $73.93$ & $73.71$ \\
		\bottomrule
	\end{tabular}
	\captionsetup{skip=0pt}
	\caption{VQA and NLVR2 evaluation results.}
	\label{tab:vqa_nlvr}
}
	\end{tabular}
\end{table*}

\begin{table*}[h!]
\begin{center}
\begin{tabular}{lc@{~~}c@{~~}cc@{~~}c@{~~}cc@{~~}c@{~~}cc@{~~}c@{~~}c}
\toprule
\multirow{3}{*}{Method}                 & \multicolumn{6}{c}{1K test set}                                               & \multicolumn{6}{c}{5K test set}                                           \\
                                        \cmidrule(lr){2-7} \cmidrule(lr){8-13}
                                        & \multicolumn{3}{c}{Text Retrieval}  & \multicolumn{3}{c}{Image Retrieval}     & \multicolumn{3}{c}{Text Retrieval}   & \multicolumn{3}{c}{Image Retrieval}  \\
                                        \cmidrule(lr){2-4}                       \cmidrule(lr){5-7}                      \cmidrule(lr){8-10}                     \cmidrule(lr){11-13}
                                        & R@1     & R@5    & R@10               & R@1    & R@5    & R@10                & R@1    & R@5    & R@10               & R@1    & R@5    & R@10   \\
\midrule
PixelBERT (R50)~\cite{abs-2004-00849}   & $77.8$  & $95.4$ & $98.2$             & $64.1$ & $91.0$ & $96.2$              & $53.4$ & $80.4$ & $88.5$             & $41.1$ & $69.7$ & $80.5$ \\             
PixelBERT (X152)~\cite{abs-2004-00849}  & $84.9$  & $97.7$ & $99.3$             & $71.6$ & $93.7$ & $97.4$              & $63.6$ & $87.5$ & $93.6$             & $50.1$ & $77.6$ & $86.2$ \\
Unicoder-VL$_\text{B}$~\cite{LiDFGJ20}  & $84.3$  & $97.3$ & $99.3$             & $69.7$ & $93.5$ & $97.2$              & $62.3$ & $87.1$ & $92.8$             & $46.7$ & $76.0$ & $85.3$ \\
OSCAR$_\text{B}$~\cite{Li0LZHZWH0WCG20} & $88.4$  & $99.1$ & $99.8$             & $75.7$ & $95.2$ & $98.3$              & $70.0$ & $91.1$ & $95.5$             & $54.0$ & $80.8$ & $88.5$ \\ 
\midrule
MiniVLM (Ours)                          &$81.1$  &$96.1$ & $99.2$               &$68.5$ &$93.0$  & $97.1$               &$58.8$  & $85.1$ & $91.7$             &$45.0$ &$74.1$   & $84.0$ \\
\bottomrule
\end{tabular}
\captionsetup{skip=0pt}
\caption{Image-Text Retrieval task evaluation results on COCO datasets. }
\label{tab:ir}
\end{center}

\end{table*}

\noindent\textbf{VQA.} The task~\cite{GoyalKSBP16} is to answer a question with natural language based on the image context, 
and we cast it as a classification problem where each class corresponds to one answer.
The representation of \texttt{[CLS]} is used to predict the answer over a shared set of $3129$ answers with a linear layer.
The model is trained with binary cross entropy loss, and the inference is to select the answer with the highest confidence. 



\noindent\textbf{Natural Language Visual Reasoning for Real (NLVR2).}
The task's input is a pair of images and a natural description, and the goal~\cite{SuhrZZZBA19} is 
to predict whether the description is true about the image pair. 
To fine-tune the network, we construct two input sequences, each containing the
concatenation of the description and one
image, and then two outputs corresponding to \texttt{[CLS]} are concatenated as the joint
representation for a binary linear classifier.

\noindent\textbf{Image-Text Retrieval.}
The task is to retrieve similar images based on the text description or vice versa. 
The key is to score the similarity of image-text pairs. 
The model is trained as a binary classification task where the input is the image region features and
the associated or mismatched text description. The transformer output corresponding to \texttt{[CLS]}
is used for binary assessment. 
The experiments are on COCO dataset, and we report top-$K$ retrieval accuracy for both $1K$ test sets and 
$5K$ test sets.

\noindent\textbf{Results.}
Table~\ref{tab:coco_caption}, ~\ref{tab:vqa_nlvr}, and~\ref{tab:ir} show the results on image captioning, VQA, NLVR2 and image-text retrieval, respectively.
As summarized in Fig.~\ref{fig:trade_off}, 
we retain $94-97\%$ of the accuracy on downstream tasks compared with the state-of-the-art model OSCAR$_\text{B}$.
In the Fig.~\ref{fig:trade_off}, captioning is measured by CIDEr score with cross entropy optimization on COCO.
Text retrieval (TR) and image retrieval (IR) are on $5K$ test set, and measured by R@10.
VQA is on the \texttt{test-std} split, and NLVR2 is on \texttt{test-P} split.
For image captioning,
the CIDEr score of OSCAR$_\text{B}$~\cite{Li0LZHZWH0WCG20} is $123.7$, while our MiniVLM achieves $119.8$ CIDEr, reaching $119.8/123.7=97\%$ accuracy. 
Compared with ~\cite{jiang2020defense}, which uses X101 to extract the grid feature, our solution achieves an even higher CIDEr ($119.8$ vs $113.8$) with much lower ($4.4$ vs $161.2$ in FLOPS) feature extraction cost as shown in Table.~\ref{tab:detector}.
On NLVR2 and image-text retrieval, our MiniVLM 
achieves higher scores than~\cite{abs-2004-00849} which uses ResNet50 as the grid feature extractor,
while both our vision and transformer modules are smaller.




\subsection{Analysis}~\label{sec:exp:analysis}

%
In this section, we provide analysis on the model architectures and pre-training methods for small VL models. Results are based on the models pretrained on the $7M$ corpus without OI-Caps-7M. All the models are evaluated through the COCO image captioning task and the VQA task.

\begin{table}[]
 \centering
 \begin{tabular}{ccccc}
 \toprule
 \multirow{2}{*}{Task}   &	\multirow{2}{*}{Init} & \multicolumn{3}{c}{Tagging Model} \\ 
                                                    \cmidrule(lr){3-5}
                            &                       & No Tag 	& TEE-0 	& TEE-3   \\
 	\midrule
    Caption                 & Text                  & $113.5$	& $117.2$   & $117.2$  	\\
 	COCO                    & Random				& $114.7$	& $116.7$   & $117.7$  	\\
 	 \midrule
VQA                         & Text                  & $67.25$	& $68.22$   & $68.72$  	\\
 	\texttt{test-dev}       & Random				& $67.05$	& $68.54$   & $68.85$  	\\
	\bottomrule

 \end{tabular}
 \caption{Impact of the tag input used in pre-training, comparing no tag with tags predicted by TEE-0, and tags predicted with higher quality by a stronger model (TEE-3). Region features are extracted with TEE-0.
    ``Text'' means the model is initialized from text pre-trained weights provided by \cite{abs-2002-10957}. 
    ``Random'' means the model is initialized from scratch.
 }
 \label{tab:tagger_cap_vqa}
\end{table}

\begin{figure}
    \centering
    \includegraphics[width=0.98\linewidth]{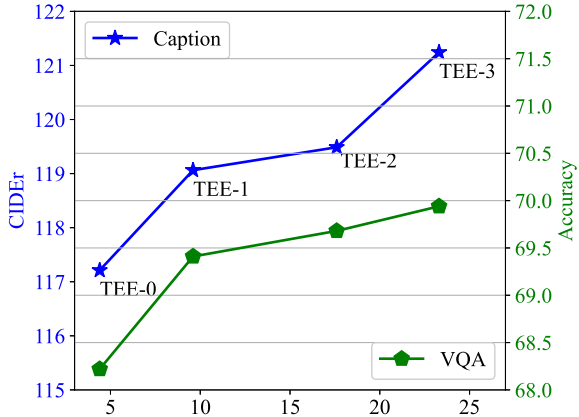}
    \caption{Impact of different backbones in TEE for COCO captioning task and VQA (\texttt{test-dev}).
    Overall, the stronger feature extractor leads to the higher score. 
    }
    \label{fig:vary_detector}
\end{figure}

\noindent\textbf{Impact of Object Tags in Pre-training.}
While Table~\ref{tab:ablation} has shown that using tags of high quality can improve the accuracy, we study 
the impact of tags under more settings in Table~\ref{tab:tagger_cap_vqa} for pre-training.
Region features are always from TEE-0 with different tagging models. 
The transformer is initialized randomly or from the pre-trained weights~\cite{abs-2002-10957} for NLP tasks. 
From the results, object tags makes large improvement ($2+$ points in caption, $1+$ in VQA) compared with the case without tags,
and high-quality tags leads to even better results. 
Small models might be more difficult to learn good representations, and thus the tag can contribute more in cross-modal alignment. 
Another observation is that random initialization gives comparable results with the text pre-trained weights.
This is similar to the findings in~\cite{TanB19}.

\noindent\textbf{Varying the backbone of TEE.}
\begin{table}[]
    \centering
    \begin{tabular}{lccc}
    \toprule
    Model                             &  Params (M)    & FLOPS (B)    & $\text{mAP}_{0.5}$    \\
    \midrule
    TEE-0                              & $7.5$          & $4.4$       & $9.9$  \\
    TEE-1                              & $10.6$         & $9.6$       & $10.6$ \\
    TEE-2                              & $12.4$         & $17.6$      & $11.3$ \\
    TEE-3                              & $17.0$         & $23.3$      & $11.5$ \\
    \bottomrule
    \end{tabular}
    \caption{
    Performance of different variants of our detectors. A larger backbone gives higher accuracy, but more cost.
    }
    \label{tab:det_var}
\end{table}
To study the impact of vision modules, we scale up TEE, ranging from TEE-0 to TEE-3 with larger sizes and better detection accuracy as shown in Table~\ref{tab:det_var}. 
As shown in Fig.~\ref{fig:vary_detector}, stronger vision module leads to better accuracy, for both caption task and VQA task.

\begin{figure}[t]
    \centering
    \includegraphics[width=0.98\linewidth]{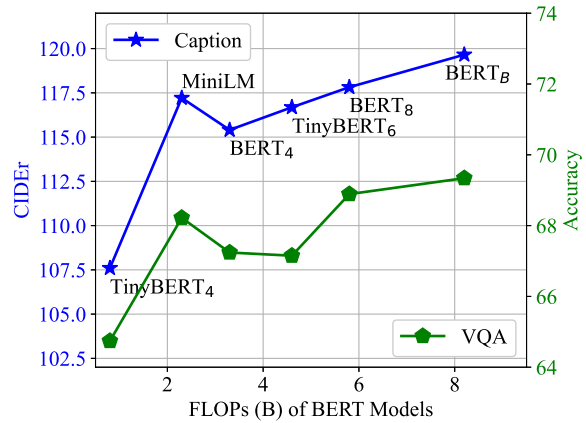}
    \caption{
    Impact of different compact BERT structures on captioning and VQA (\texttt{test-dev}). 
   	}
    \label{fig:bert}
\end{figure}

\noindent\textbf{Impact of Compact BERT Structures.}
Fig.~\ref{fig:bert} shows the experimental results on speed-accuracy trade-off for models with different transformer modules listed
in Table~\ref{tab:bert_cost}.
Among the structures, MiniLM achieves a better trade-off between speed and accuracy. 
This shows that a ``thinner'' version of BERT could make better trade-off than the ``shallower'' version for VL tasks.





\section{Conclusion}
In this paper, we have proposed a compact solution, MiniVLM, for vision-language (VL) tasks,
which is smaller and faster, and thus can be deployed in real-world
applications on resource-constrained devices. 
For the vision module,
we design the Two-stage Efficient feature Extractor (TEE), to significantly save computation by simplifying the region head and replacing regular convolutional layers with pointwise and depthwise convolution layers.
To improve the small-model pre-training, we leverage large models and large-scale dataset.
We fine-tune the pre-trained model on various downstream VL tasks, and show that MiniVLM can retain $94-97\%$ of the accuracy with $27\%$ parameters and $1\%$ FLOPS compared to the state-of-the-art VL model.

{\small
\bibliographystyle{ieee_fullname}
\bibliography{egbib}

\begin{thebibliography}{10}\itemsep=-1pt

\bibitem{AndersonFJG16}
Peter Anderson, Basura Fernando, Mark Johnson, and Stephen Gould.
\newblock {SPICE:} semantic propositional image caption evaluation.
\newblock In {\em Computer Vision - {ECCV} 2016 - 14th European Conference,
  Amsterdam, The Netherlands, October 11-14, 2016, Proceedings, Part {V}},
  volume 9909 of {\em Lecture Notes in Computer Science}, pages 382--398.
  Springer, 2016.

\bibitem{00010BT0GZ18}
Peter Anderson, Xiaodong He, Chris Buehler, Damien Teney, Mark Johnson, Stephen
  Gould, and Lei Zhang.
\newblock Bottom-up and top-down attention for image captioning and visual
  question answering.
\newblock In {\em 2018 {IEEE} Conference on Computer Vision and Pattern
  Recognition, {CVPR} 2018, Salt Lake City, UT, USA, June 18-22, 2018}, pages
  6077--6086. {IEEE} Computer Society, 2018.

\bibitem{CarionMSUKZ20}
Nicolas Carion, Francisco Massa, Gabriel Synnaeve, Nicolas Usunier, Alexander
  Kirillov, and Sergey Zagoruyko.
\newblock End-to-end object detection with transformers.
\newblock In {\em Computer Vision - {ECCV} 2020 - 16th European Conference,
  Glasgow, UK, August 23-28, 2020, Proceedings, Part {I}}, volume 12346 of {\em
  Lecture Notes in Computer Science}, pages 213--229. Springer, 2020.

\bibitem{ChenLYK0G0020}
Yen{-}Chun Chen, Linjie Li, Licheng Yu, Ahmed~El Kholy, Faisal Ahmed, Zhe Gan,
  Yu Cheng, and Jingjing Liu.
\newblock {UNITER:} universal image-text representation learning.
\newblock In {\em Computer Vision - {ECCV} 2020 - 16th European Conference,
  Glasgow, UK, August 23-28, 2020, Proceedings, Part {XXX}}, volume 12375 of
  {\em Lecture Notes in Computer Science}, pages 104--120. Springer, 2020.

\bibitem{DenkowskiL14}
Michael~J. Denkowski and Alon Lavie.
\newblock Meteor universal: Language specific translation evaluation for any
  target language.
\newblock In {\em Proceedings of the Ninth Workshop on Statistical Machine
  Translation, WMT@ACL 2014, June 26-27, 2014, Baltimore, Maryland, {USA}},
  pages 376--380. The Association for Computer Linguistics, 2014.

\bibitem{DevlinCLT19}
Jacob Devlin, Ming{-}Wei Chang, Kenton Lee, and Kristina Toutanova.
\newblock {BERT:} pre-training of deep bidirectional transformers for language
  understanding.
\newblock In {\em Proceedings of the 2019 Conference of the North American
  Chapter of the Association for Computational Linguistics: Human Language
  Technologies, {NAACL-HLT} 2019, Minneapolis, MN, USA, June 2-7, 2019, Volume
  1 (Long and Short Papers)}, pages 4171--4186. Association for Computational
  Linguistics, 2019.

\bibitem{GhiasiLL19}
Golnaz Ghiasi, Tsung{-}Yi Lin, and Quoc~V. Le.
\newblock {NAS-FPN:} learning scalable feature pyramid architecture for object
  detection.
\newblock In {\em {IEEE} Conference on Computer Vision and Pattern Recognition,
  {CVPR} 2019, Long Beach, CA, USA, June 16-20, 2019}, pages 7036--7045.
  Computer Vision Foundation / {IEEE}, 2019.

\bibitem{GoyalKSBP16}
Yash Goyal, Tejas Khot, Douglas Summers{-}Stay, Dhruv Batra, and Devi Parikh.
\newblock Making the {V} in {VQA} matter: Elevating the role of image
  understanding in visual question answering.
\newblock {\em CoRR}, abs/1612.00837, 2016.

\bibitem{HeGDG20}
Kaiming He, Georgia Gkioxari, Piotr Doll{\'{a}}r, and Ross~B. Girshick.
\newblock Mask {R-CNN}.
\newblock {\em {IEEE} Trans. Pattern Anal. Mach. Intell.}, 42(2):386--397,
  2020.

\bibitem{HeZRS16}
Kaiming He, Xiangyu Zhang, Shaoqing Ren, and Jian Sun.
\newblock Deep residual learning for image recognition.
\newblock In {\em 2016 {IEEE} Conference on Computer Vision and Pattern
  Recognition, {CVPR} 2016, Las Vegas, NV, USA, June 27-30, 2016}, pages
  770--778. {IEEE} Computer Society, 2016.

\bibitem{HintonVD15}
Geoffrey~E. Hinton, Oriol Vinyals, and Jeffrey Dean.
\newblock Distilling the knowledge in a neural network.
\newblock {\em CoRR}, abs/1503.02531, 2015.

\bibitem{sangBS17}
Jan~Hendrik Hosang, Rodrigo Benenson, and Bernt Schiele.
\newblock Learning non-maximum suppression.
\newblock In {\em 2017 {IEEE} Conference on Computer Vision and Pattern
  Recognition, {CVPR} 2017, Honolulu, HI, USA, July 21-26, 2017}, pages
  6469--6477. {IEEE} Computer Society, 2017.

\bibitem{abs-2009-13682}
Xiaowei Hu, Xi Yin, Kevin Lin, Lijuan Wang, Lei Zhang, Jianfeng Gao, and
  Zicheng Liu.
\newblock {VIVO:} surpassing human performance in novel object captioning with
  visual vocabulary pre-training.
\newblock {\em CoRR}, abs/2009.13682, 2020.

\bibitem{huang2019attention}
Lun Huang, Wenmin Wang, Jie Chen, and Xiao-Yong Wei.
\newblock Attention on attention for image captioning.
\newblock In {\em International Conference on Computer Vision}, 2019.

\bibitem{abs-2004-00849}
Zhicheng Huang, Zhaoyang Zeng, Bei Liu, Dongmei Fu, and Jianlong Fu.
\newblock Pixel-bert: Aligning image pixels with text by deep multi-modal
  transformers.
\newblock {\em CoRR}, abs/2004.00849, 2020.

\bibitem{HudsonM19}
Drew~A. Hudson and Christopher~D. Manning.
\newblock {GQA:} {A} new dataset for real-world visual reasoning and
  compositional question answering.
\newblock In {\em {IEEE} Conference on Computer Vision and Pattern Recognition,
  {CVPR} 2019, Long Beach, CA, USA, June 16-20, 2019}, pages 6700--6709.
  Computer Vision Foundation / {IEEE}, 2019.

\bibitem{jiang2020defense}
Huaizu Jiang, Ishan Misra, Marcus Rohrbach, Erik Learned-Miller, and Xinlei
  Chen.
\newblock In defense of grid features for visual question answering.
\newblock In {\em Proceedings of the IEEE/CVF Conference on Computer Vision and
  Pattern Recognition}, pages 10267--10276, 2020.

\bibitem{abs-1909-10351}
Xiaoqi Jiao, Yichun Yin, Lifeng Shang, Xin Jiang, Xiao Chen, Linlin Li, Fang
  Wang, and Qun Liu.
\newblock Tinybert: Distilling {BERT} for natural language understanding.
\newblock {\em CoRR}, abs/1909.10351, 2019.

\bibitem{KarpathyL15}
Andrej Karpathy and Fei{-}Fei Li.
\newblock Deep visual-semantic alignments for generating image descriptions.
\newblock In {\em {IEEE} Conference on Computer Vision and Pattern Recognition,
  {CVPR} 2015, Boston, MA, USA, June 7-12, 2015}, pages 3128--3137. {IEEE}
  Computer Society, 2015.

\bibitem{KrishnaZGJHKCKL16}
Ranjay Krishna, Yuke Zhu, Oliver Groth, Justin Johnson, Kenji Hata, Joshua
  Kravitz, Stephanie Chen, Yannis Kalantidis, Li{-}Jia Li, David~A. Shamma,
  Michael~S. Bernstein, and Fei{-}Fei Li.
\newblock Visual genome: Connecting language and vision using crowdsourced
  dense image annotations.
\newblock {\em CoRR}, abs/1602.07332, 2016.

\bibitem{OpenImages}
Alina Kuznetsova, Hassan Rom, Neil Alldrin, Jasper Uijlings, Ivan Krasin, Jordi
  Pont-Tuset, Shahab Kamali, Stefan Popov, Matteo Malloci, Alexander
  Kolesnikov, Tom Duerig, and Vittorio Ferrari.
\newblock The open images dataset v4: Unified image classification, object
  detection, and visual relationship detection at scale.
\newblock {\em IJCV}, 2020.

\bibitem{abs-1909-11942}
Zhenzhong Lan, Mingda Chen, Sebastian Goodman, Kevin Gimpel, Piyush Sharma, and
  Radu Soricut.
\newblock {ALBERT:} {A} lite {BERT} for self-supervised learning of language
  representations.
\newblock {\em CoRR}, abs/1909.11942, 2019.

\bibitem{abs-1904-08900}
Hei Law, Yun Teng, Olga Russakovsky, and Jia Deng.
\newblock Cornernet-lite: Efficient keypoint based object detection.
\newblock {\em CoRR}, abs/1904.08900, 2019.

\bibitem{LiDFGJ20}
Gen Li, Nan Duan, Yuejian Fang, Ming Gong, and Daxin Jiang.
\newblock Unicoder-vl: {A} universal encoder for vision and language by
  cross-modal pre-training.
\newblock In {\em The Thirty-Fourth {AAAI} Conference on Artificial
  Intelligence, {AAAI} 2020, The Thirty-Second Innovative Applications of
  Artificial Intelligence Conference, {IAAI} 2020, The Tenth {AAAI} Symposium
  on Educational Advances in Artificial Intelligence, {EAAI} 2020, New York,
  NY, USA, February 7-12, 2020}, pages 11336--11344. {AAAI} Press, 2020.

\bibitem{abs-1908-03557}
Liunian~Harold Li, Mark Yatskar, Da Yin, Cho{-}Jui Hsieh, and Kai{-}Wei Chang.
\newblock Visualbert: {A} simple and performant baseline for vision and
  language.
\newblock {\em CoRR}, abs/1908.03557, 2019.

\bibitem{Li0LZHZWH0WCG20}
Xiujun Li, Xi Yin, Chunyuan Li, Pengchuan Zhang, Xiaowei Hu, Lei Zhang, Lijuan
  Wang, Houdong Hu, Li Dong, Furu Wei, Yejin Choi, and Jianfeng Gao.
\newblock Oscar: Object-semantics aligned pre-training for vision-language
  tasks.
\newblock In {\em Computer Vision - {ECCV} 2020 - 16th European Conference,
  Glasgow, UK, August 23-28, 2020, Proceedings, Part {XXX}}, volume 12375 of
  {\em Lecture Notes in Computer Science}, pages 121--137. Springer, 2020.

\bibitem{abs-1711-07264}
Zeming Li, Chao Peng, Gang Yu, Xiangyu Zhang, Yangdong Deng, and Jian Sun.
\newblock Light-head {R-CNN:} in defense of two-stage object detector.
\newblock {\em CoRR}, abs/1711.07264, 2017.

\bibitem{lin2020interbert}
Junyang Lin, An Yang, Yichang Zhang, Jie Liu, Jingren Zhou, and Hongxia Yang.
\newblock Interbert: Vision-and-language interaction for multi-modal
  pretraining.
\newblock {\em arXiv preprint arXiv:2003.13198}, 2020.

\bibitem{LinDGHHB17}
Tsung{-}Yi Lin, Piotr Doll{\'{a}}r, Ross~B. Girshick, Kaiming He, Bharath
  Hariharan, and Serge~J. Belongie.
\newblock Feature pyramid networks for object detection.
\newblock In {\em 2017 {IEEE} Conference on Computer Vision and Pattern
  Recognition, {CVPR} 2017, Honolulu, HI, USA, July 21-26, 2017}, pages
  936--944. {IEEE} Computer Society, 2017.

\bibitem{LinGGHD20}
Tsung{-}Yi Lin, Priya Goyal, Ross~B. Girshick, Kaiming He, and Piotr
  Doll{\'{a}}r.
\newblock Focal loss for dense object detection.
\newblock {\em {IEEE} Trans. Pattern Anal. Mach. Intell.}, 42(2):318--327,
  2020.

\bibitem{LinMBHPRDZ14}
Tsung{-}Yi Lin, Michael Maire, Serge~J. Belongie, Lubomir~D. Bourdev, Ross~B.
  Girshick, James Hays, Pietro Perona, Deva Ramanan, Piotr Doll{\'{a}}r, and
  C.~Lawrence Zitnick.
\newblock Microsoft {COCO:} common objects in context.
\newblock {\em CoRR}, abs/1405.0312, 2014.

\bibitem{LiuAESRFB16}
Wei Liu, Dragomir Anguelov, Dumitru Erhan, Christian Szegedy, Scott~E. Reed,
  Cheng{-}Yang Fu, and Alexander~C. Berg.
\newblock {SSD:} single shot multibox detector.
\newblock In {\em Computer Vision - {ECCV} 2016 - 14th European Conference,
  Amsterdam, The Netherlands, October 11-14, 2016, Proceedings, Part {I}},
  volume 9905 of {\em Lecture Notes in Computer Science}, pages 21--37.
  Springer, 2016.

\bibitem{LoshchilovH17}
Ilya Loshchilov and Frank Hutter.
\newblock {SGDR:} stochastic gradient descent with warm restarts.
\newblock In {\em 5th International Conference on Learning Representations,
  {ICLR} 2017, Toulon, France, April 24-26, 2017, Conference Track
  Proceedings}. OpenReview.net, 2017.

\bibitem{LoshchilovH19}
Ilya Loshchilov and Frank Hutter.
\newblock Decoupled weight decay regularization.
\newblock In {\em 7th International Conference on Learning Representations,
  {ICLR} 2019, New Orleans, LA, USA, May 6-9, 2019}. OpenReview.net, 2019.

\bibitem{LuBPL19}
Jiasen Lu, Dhruv Batra, Devi Parikh, and Stefan Lee.
\newblock Vilbert: Pretraining task-agnostic visiolinguistic representations
  for vision-and-language tasks.
\newblock In {\em Advances in Neural Information Processing Systems 32: Annual
  Conference on Neural Information Processing Systems 2019, NeurIPS 2019, 8-14
  December 2019, Vancouver, BC, Canada}, pages 13--23, 2019.

\bibitem{MehtaRSH19}
Sachin Mehta, Mohammad Rastegari, Linda~G. Shapiro, and Hannaneh Hajishirzi.
\newblock Espnetv2: {A} light-weight, power efficient, and general purpose
  convolutional neural network.
\newblock In {\em {IEEE} Conference on Computer Vision and Pattern Recognition,
  {CVPR} 2019, Long Beach, CA, USA, June 16-20, 2019}, pages 9190--9200.
  Computer Vision Foundation / {IEEE}, 2019.

\bibitem{OrdonezKB11}
Vicente Ordonez, Girish Kulkarni, and Tamara~L. Berg.
\newblock Im2text: Describing images using 1 million captioned photographs.
\newblock In {\em Advances in Neural Information Processing Systems 24: 25th
  Annual Conference on Neural Information Processing Systems 2011. Proceedings
  of a meeting held 12-14 December 2011, Granada, Spain}, pages 1143--1151,
  2011.

\bibitem{PapineniRWZ02}
Kishore Papineni, Salim Roukos, Todd Ward, and Wei{-}Jing Zhu.
\newblock Bleu: a method for automatic evaluation of machine translation.
\newblock In {\em Proceedings of the 40th Annual Meeting of the Association for
  Computational Linguistics, July 6-12, 2002, Philadelphia, PA, {USA}}, pages
  311--318. {ACL}, 2002.

\bibitem{RadosavovicDGGH18}
Ilija Radosavovic, Piotr Doll{\'{a}}r, Ross~B. Girshick, Georgia Gkioxari, and
  Kaiming He.
\newblock Data distillation: Towards omni-supervised learning.
\newblock In {\em 2018 {IEEE} Conference on Computer Vision and Pattern
  Recognition, {CVPR} 2018, Salt Lake City, UT, USA, June 18-22, 2018}, pages
  4119--4128. {IEEE} Computer Society, 2018.

\bibitem{RedmonDGF16}
Joseph Redmon, Santosh~Kumar Divvala, Ross~B. Girshick, and Ali Farhadi.
\newblock You only look once: Unified, real-time object detection.
\newblock In {\em 2016 {IEEE} Conference on Computer Vision and Pattern
  Recognition, {CVPR} 2016, Las Vegas, NV, USA, June 27-30, 2016}, pages
  779--788. {IEEE} Computer Society, 2016.

\bibitem{RenHGS15}
Shaoqing Ren, Kaiming He, Ross~B. Girshick, and Jian Sun.
\newblock Faster {R-CNN:} towards real-time object detection with region
  proposal networks.
\newblock In {\em Advances in Neural Information Processing Systems 28: Annual
  Conference on Neural Information Processing Systems 2015, December 7-12,
  2015, Montreal, Quebec, Canada}, pages 91--99, 2015.

\bibitem{RussakovskyDSKS15}
Olga Russakovsky, Jia Deng, Hao Su, Jonathan Krause, Sanjeev Satheesh, Sean Ma,
  Zhiheng Huang, Andrej Karpathy, Aditya Khosla, Michael~S. Bernstein,
  Alexander~C. Berg, and Fei{-}Fei Li.
\newblock Imagenet large scale visual recognition challenge.
\newblock {\em Int. J. Comput. Vis.}, 115(3):211--252, 2015.

\bibitem{abs-1910-01108}
Victor Sanh, Lysandre Debut, Julien Chaumond, and Thomas Wolf.
\newblock Distilbert, a distilled version of {BERT:} smaller, faster, cheaper
  and lighter.
\newblock {\em CoRR}, abs/1910.01108, 2019.

\bibitem{0005LZPYZLS19}
Shuai Shao, Zeming Li, Tianyuan Zhang, Chao Peng, Gang Yu, Xiangyu Zhang, Jing
  Li, and Jian Sun.
\newblock Objects365: {A} large-scale, high-quality dataset for object
  detection.
\newblock In {\em 2019 {IEEE/CVF} International Conference on Computer Vision,
  {ICCV} 2019, Seoul, Korea (South), October 27 - November 2, 2019}, pages
  8429--8438. {IEEE}, 2019.

\bibitem{SoricutDSG18}
Piyush Sharma, Nan Ding, Sebastian Goodman, and Radu Soricut.
\newblock Conceptual captions: {A} cleaned, hypernymed, image alt-text dataset
  for automatic image captioning.
\newblock In {\em Proceedings of the 56th Annual Meeting of the Association for
  Computational Linguistics, {ACL} 2018, Melbourne, Australia, July 15-20,
  2018, Volume 1: Long Papers}, pages 2556--2565. Association for Computational
  Linguistics, 2018.

\bibitem{SuZCLLWD20}
Weijie Su, Xizhou Zhu, Yue Cao, Bin Li, Lewei Lu, Furu Wei, and Jifeng Dai.
\newblock {VL-BERT:} pre-training of generic visual-linguistic representations.
\newblock In {\em 8th International Conference on Learning Representations,
  {ICLR} 2020, Addis Ababa, Ethiopia, April 26-30, 2020}. OpenReview.net, 2020.

\bibitem{SuhrZZZBA19}
Alane Suhr, Stephanie Zhou, Ally Zhang, Iris Zhang, Huajun Bai, and Yoav Artzi.
\newblock A corpus for reasoning about natural language grounded in
  photographs.
\newblock In {\em Proceedings of the 57th Conference of the Association for
  Computational Linguistics, {ACL} 2019, Florence, Italy, July 28- August 2,
  2019, Volume 1: Long Papers}, pages 6418--6428. Association for Computational
  Linguistics, 2019.

\bibitem{peize2020sparse}
Peize Sun, Rufeng Zhang, Yi Jiang, Tao Kong, Chenfeng Xu, Wei Zhan, Masayoshi
  Tomizuka, Lei Li, Zehuan Yuan, Changhu Wang, and Ping Luo.
\newblock {SparseR-CNN}: End-to-end object detection with learnable proposals.
\newblock {\em arXiv preprint arXiv:2011.12450}, 2020.

\bibitem{SunYSLYZ20}
Zhiqing Sun, Hongkun Yu, Xiaodan Song, Renjie Liu, Yiming Yang, and Denny Zhou.
\newblock Mobilebert: a compact task-agnostic {BERT} for resource-limited
  devices.
\newblock In {\em Proceedings of the 58th Annual Meeting of the Association for
  Computational Linguistics, {ACL} 2020, Online, July 5-10, 2020}, pages
  2158--2170. Association for Computational Linguistics, 2020.

\bibitem{TanB19}
Hao Tan and Mohit Bansal.
\newblock {LXMERT:} learning cross-modality encoder representations from
  transformers.
\newblock In {\em Proceedings of the 2019 Conference on Empirical Methods in
  Natural Language Processing and the 9th International Joint Conference on
  Natural Language Processing, {EMNLP-IJCNLP} 2019, Hong Kong, China, November
  3-7, 2019}, pages 5099--5110. Association for Computational Linguistics,
  2019.

\bibitem{TanL19}
Mingxing Tan and Quoc~V. Le.
\newblock Efficientnet: Rethinking model scaling for convolutional neural
  networks.
\newblock In {\em Proceedings of the 36th International Conference on Machine
  Learning, {ICML} 2019, 9-15 June 2019, Long Beach, California, {USA}},
  volume~97 of {\em Proceedings of Machine Learning Research}, pages
  6105--6114. {PMLR}, 2019.

\bibitem{TanPL20}
Mingxing Tan, Ruoming Pang, and Quoc~V. Le.
\newblock Efficientdet: Scalable and efficient object detection.
\newblock In {\em 2020 {IEEE/CVF} Conference on Computer Vision and Pattern
  Recognition, {CVPR} 2020, Seattle, WA, USA, June 13-19, 2020}, pages
  10778--10787. {IEEE}, 2020.

\bibitem{abs-2006-09214}
Zhi Tian, Chunhua Shen, Hao Chen, and Tong He.
\newblock {FCOS:} {A} simple and strong anchor-free object detector.
\newblock {\em CoRR}, abs/2006.09214, 2020.

\bibitem{VaswaniSPUJGKP17}
Ashish Vaswani, Noam Shazeer, Niki Parmar, Jakob Uszkoreit, Llion Jones,
  Aidan~N. Gomez, Lukasz Kaiser, and Illia Polosukhin.
\newblock Attention is all you need.
\newblock In {\em Advances in Neural Information Processing Systems 30: Annual
  Conference on Neural Information Processing Systems 2017, 4-9 December 2017,
  Long Beach, CA, {USA}}, pages 5998--6008, 2017.

\bibitem{VedantamZP15}
Ramakrishna Vedantam, C.~Lawrence Zitnick, and Devi Parikh.
\newblock Cider: Consensus-based image description evaluation.
\newblock In {\em {IEEE} Conference on Computer Vision and Pattern Recognition,
  {CVPR} 2015, Boston, MA, USA, June 7-12, 2015}, pages 4566--4575. {IEEE}
  Computer Society, 2015.

\bibitem{abs-2005-11426}
Jianfeng Wang, Xi Yin, Lijuan Wang, and Lei Zhang.
\newblock Hashing-based non-maximum suppression for crowded object detection.
\newblock {\em CoRR}, abs/2005.11426, 2020.

\bibitem{Wang0AL18}
Robert~J. Wang, Xiang Li, Shuang Ao, and Charles~X. Ling.
\newblock Pelee: {A} real-time object detection system on mobile devices.
\newblock In {\em 6th International Conference on Learning Representations,
  {ICLR} 2018, Vancouver, BC, Canada, April 30 - May 3, 2018, Workshop Track
  Proceedings}. OpenReview.net, 2018.

\bibitem{WangACKP019}
Tiancai Wang, Rao~Muhammad Anwer, Hisham Cholakkal, Fahad~Shahbaz Khan, Yanwei
  Pang, and Ling Shao.
\newblock Learning rich features at high-speed for single-shot object
  detection.
\newblock In {\em 2019 {IEEE/CVF} International Conference on Computer Vision,
  {ICCV} 2019, Seoul, Korea (South), October 27 - November 2, 2019}, pages
  1971--1980. {IEEE}, 2019.

\bibitem{abs-2002-10957}
Wenhui Wang, Furu Wei, Li Dong, Hangbo Bao, Nan Yang, and Ming Zhou.
\newblock Minilm: Deep self-attention distillation for task-agnostic
  compression of pre-trained transformers.
\newblock {\em CoRR}, abs/2002.10957, 2020.

\bibitem{0008CYLWL020}
Yue Wu, Yinpeng Chen, Lu Yuan, Zicheng Liu, Lijuan Wang, Hongzhi Li, and Yun
  Fu.
\newblock Rethinking classification and localization for object detection.
\newblock In {\em 2020 {IEEE/CVF} Conference on Computer Vision and Pattern
  Recognition, {CVPR} 2020, Seattle, WA, USA, June 13-19, 2020}, pages
  10183--10192. {IEEE}, 2020.

\bibitem{XieLHL20}
Qizhe Xie, Minh{-}Thang Luong, Eduard~H. Hovy, and Quoc~V. Le.
\newblock Self-training with noisy student improves imagenet classification.
\newblock In {\em 2020 {IEEE/CVF} Conference on Computer Vision and Pattern
  Recognition, {CVPR} 2020, Seattle, WA, USA, June 13-19, 2020}, pages
  10684--10695. {IEEE}, 2020.

\bibitem{abs-1905-00546}
I.~Zeki Yalniz, Herv{\'{e}} J{\'{e}}gou, Kan Chen, Manohar Paluri, and Dhruv
  Mahajan.
\newblock Billion-scale semi-supervised learning for image classification.
\newblock {\em CoRR}, abs/1905.00546, 2019.

\bibitem{YoungLHH14}
Peter Young, Alice Lai, Micah Hodosh, and Julia Hockenmaier.
\newblock From image descriptions to visual denotations: New similarity metrics
  for semantic inference over event descriptions.
\newblock {\em Trans. Assoc. Comput. Linguistics}, 2:67--78, 2014.

\bibitem{abs-2006-16934}
Fei Yu, Jiji Tang, Weichong Yin, Yu Sun, Hao Tian, Hua Wu, and Haifeng Wang.
\newblock Ernie-vil: Knowledge enhanced vision-language representations through
  scene graph.
\newblock {\em CoRR}, abs/2006.16934, 2020.

\bibitem{ZhangCYLL20}
Shifeng Zhang, Cheng Chi, Yongqiang Yao, Zhen Lei, and Stan~Z. Li.
\newblock Bridging the gap between anchor-based and anchor-free detection via
  adaptive training sample selection.
\newblock In {\em 2020 {IEEE/CVF} Conference on Computer Vision and Pattern
  Recognition, {CVPR} 2020, Seattle, WA, USA, June 13-19, 2020}, pages
  9756--9765. {IEEE}, 2020.

\bibitem{ZhouPZHCG20}
Luowei Zhou, Hamid Palangi, Lei Zhang, Houdong Hu, Jason~J. Corso, and Jianfeng
  Gao.
\newblock Unified vision-language pre-training for image captioning and {VQA}.
\newblock In {\em The Thirty-Fourth {AAAI} Conference on Artificial
  Intelligence, {AAAI} 2020, The Thirty-Second Innovative Applications of
  Artificial Intelligence Conference, {IAAI} 2020, The Tenth {AAAI} Symposium
  on Educational Advances in Artificial Intelligence, {EAAI} 2020, New York,
  NY, USA, February 7-12, 2020}, pages 13041--13049. {AAAI} Press, 2020.

\end{thebibliography}
}

\end{document}